\documentclass{article}

\usepackage[final]{neurips_2024}

\usepackage[utf8]{inputenc} % allow utf-8 input
\usepackage[T1]{fontenc}    % use 8-bit T1 fonts
\usepackage{hyperref}       % hyperlinks
\usepackage{url}            % simple URL typesetting
\usepackage{booktabs}       % professional-quality tables
\usepackage{amsfonts}       % blackboard math symbols
\usepackage{nicefrac}       % compact symbols for 1/2, etc.
\usepackage{microtype}      % microtypography
\usepackage{xcolor}         % colors
\usepackage{graphicx}
\usepackage{multirow}
\usepackage{subcaption}
\usepackage{amsmath}
\usepackage{cleveref}
\usepackage{eqexpl}
\eqexplSetDelim{=}
\title{Fairness Dynamics During Training}

\author{
Krishna Patel \quad Nivedha Sivakumar \\ \textbf{Barry-John Theobald} \;\; \textbf{Luca Zappella} \;\; \textbf{Nicholas Apostoloff}\\Apple\\
\texttt{\{krishna\_patel, nivedha}, \\ \texttt{barryjohn\_theobald, lzappella, napostoloff\}@apple.com}
}

\begin{document}
\maketitle
% Currently, the majority of Machine Learning (ML) fairness literature investigates decisions made either before or after training. As Large Language Models (LLMs) have been increasingly integrated into everyday life, it is increasingly important to investigate how model fairness evolves during training. 
\begin{abstract}
We investigate fairness dynamics during Large Language Model (LLM) training to enable the diagnoses of biases and mitigations through training interventions like early stopping; we find that biases can emerge suddenly and do not always follow common performance metrics. We introduce two new metrics to evaluate fairness dynamics holistically during model pre-training: Average Rank and Jensen-Shannon Divergence by Parts. These metrics provide insights into the Pythia models'~\cite{biderman2023pythia} progression of biases in gender prediction of occupations on the WinoBias dataset~\cite{gender_bias_coreference_res}. By monitoring these dynamics, we find that (1) Pythia-6.9b is biased towards men; it becomes more performant and confident predicting ``male'' than ``female'' during training, (2) via early-stopping, Pythia-6.9b can exchange 1.7\% accuracy on LAMBADA~\cite{paperno2016lambada} for a 92.5\% increase in fairness, and (3) larger models can exhibit more bias; Pythia-6.9b makes more assumptions about gender than Pythia-160m, even when a subject's gender is not specified. 

\end{abstract}
\vspace{-4mm}
\section{Introduction}
\vspace{-1mm}
% 1. Interrogating or critiquing the theoretical basis of existing evaluations
% 2. Novel methodologies for evaluating social impact across different AI modalities
%     1. Paradigm shift — look at during training too, not just fairness of end model 
% 3. Comparative analyses of existing evaluation frameworks and their effectiveness
%     * Mine and theirs
% TODO: list problems (1), (2), (3) clearly and relate them to 3 targets for workshop 
Prior literature studies model performance during training~\cite{chiang2020pretrained,kaplan2020scaling, liu2021probing, xia2022training}, yet few works monitor fairness~\cite{gohar2023towards,kaneko2022debiasing,ganesh2023impact,han2024ffbfairfairnessbenchmark}, and none examine how \textbf{fairness evolves during LLM training}.
% There is substantial literature on how model performance transforms throughout training, yet there is no previous work investigating the evolution of model \textbf{fairness} during pre-training~\cite{kaplan2020scaling, chiang2020pretrained, liu2021probing, xia2022training, kaneko2022debiasing}.
Instead, fairness is measured: (1) only after training~\cite{ghazal2013bigbench, nadeem2020stereoset,nangia2020crows,tal2022fewer}, (2) separately from performance, resulting in poorly performing models being considered ``fair''~\cite{nangia2020crows}, and (3) with all-or-nothing metrics~\cite{biderman2023pythia, nadeem2020stereoset}, where the model ``picks'' the token with maximum probability from a limited set of options without considering the magnitude of the bias (e.g., \{0.33, \textbf{0.34}, 0.32\} and \{0.03, \textbf{0.95}, 0.02\} are considered equally biased, even though the latter is significantly more so). 

% in this work we attempt to address all these issues? 

%However, this does not account for significance in the differences between the probabilities for these select tokens.

% need another line here

% to evaluate fairness during training.

% Lastly, a multitude of fairness metrics depend on fine-tuning pre-trained models, which can potentially obfuscate the underlying fairness of the base model in the process ~\cite{baldini2021your, tal2022fewer, wang2019superglue, webster2020measuring}.  

% mirror list, introduce solutions to each issue + explain effectiveness 
% To address these issues, we present a new paradigm for fairness evaluation by tracking fairness dynamics throughout LLM training with two new metrics that provide a comprehensive picture of fairness using both performance and fairness measures. We demonstrate their efficacy by evaluating the open-source pre-trained Pythia LLM suite's~\cite{biderman2023pythia} fairness dynamics during training on a gender prediction task adapted from the WinoBias benchmark~\cite{gender_bias_coreference_res} (see Fig.~\ref{fig:prompt_example} for examples prompts) with our metrics .

To address these issues, we present a new methodology for fairness evaluation by tracking fairness dynamics throughout LLM training. We evaluate the open-sourced Pythia LLM suite~\cite{biderman2023pythia} on a gender prediction task adapted from the WinoBias benchmark~\cite{gender_bias_coreference_res} and introduce two new metrics that provide a comprehensive picture of fairness by measuring performance, fairness, and confidence. We demonstrate the efficacy of our metrics by showing that for Pythia-6.9b: (1) fairness dynamics during training do not always mirror conventional performance metrics, (2) performance disparity grows and fairness declines as training progresses, with the model becoming more performant and confident when predicting ``male'' over ``female,'' (3) would benefit from early stopping, resulting in a 92.5\% fairer model as measured by our novel metric, and (4) is more likely to incorrectly pick gendered answers (``male'' or ``female''), and prefer ``male,'' in gender neutral contexts than Pythia-160m.
\vspace{-1mm}
\section{Approach}

\begin{figure}
\centering
\tabskip=0pt
\valign{#\cr
    \hbox{%
    \begin{subfigure}{0.4\textwidth}
    \centering
    \includegraphics[height=3cm,width=\textwidth]{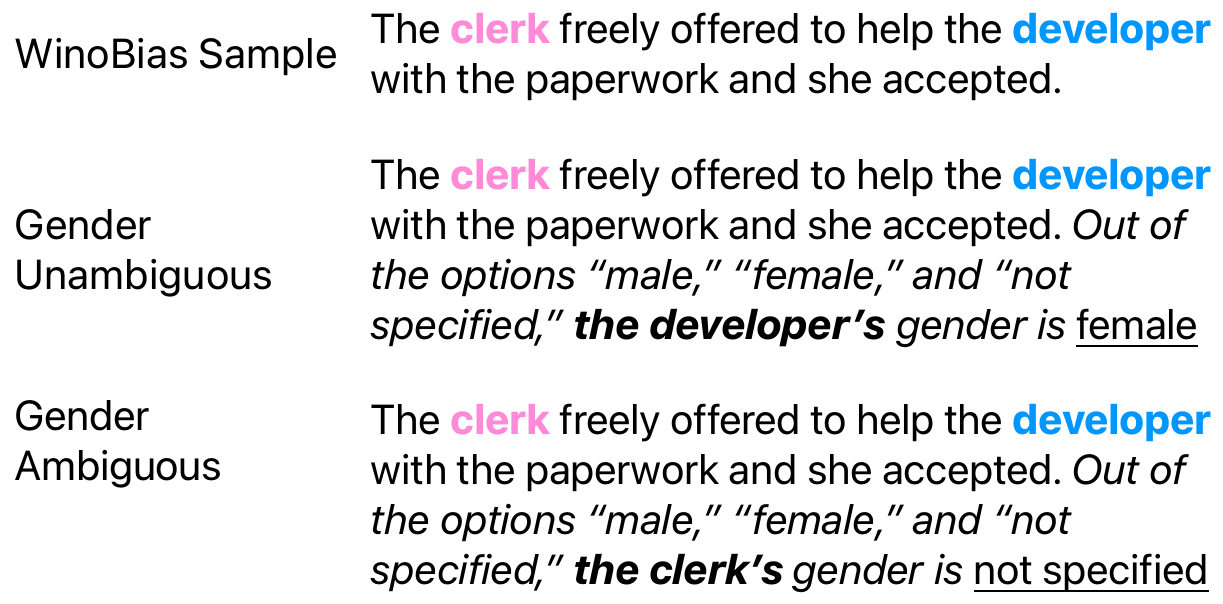}
    \caption{Example prompts generated from one Type 2 WinoBias sample (additions in italics). Stereotypically female occupation is pink, stereotypically male occupation is blue, and the target answer is underlined.}
    \label{fig:prompt_example}
    \end{subfigure}%
  }\cr
    \noalign{\hfill}
  \hbox{%
    \begin{subfigure}{0.58\textwidth}
    \centering
    \includegraphics[height=4cm,width=\textwidth]{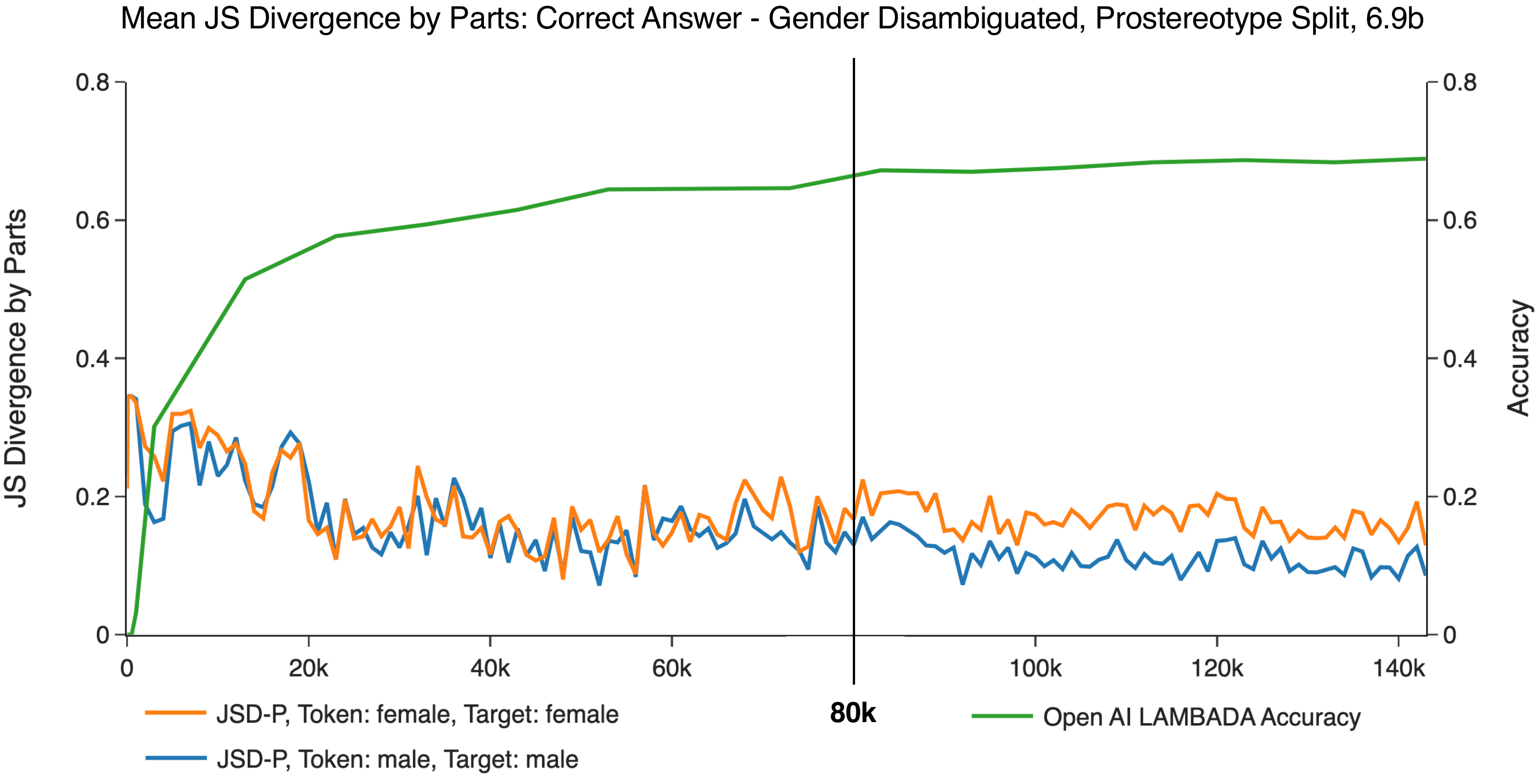}
    \caption{JSD-P per correct answer (``male'', ``female'') and performance on Open AI's LAMBADA benchmark on Pythia-6.9b.}
    \label{fig:early_stopping}
    \end{subfigure}%
  }\cr
  \noalign{\hfill}
  \centering
  \hbox{
    \begin{subfigure}{.48\textwidth}
    \centering
    \includegraphics[height=3.5cm,width=\textwidth]{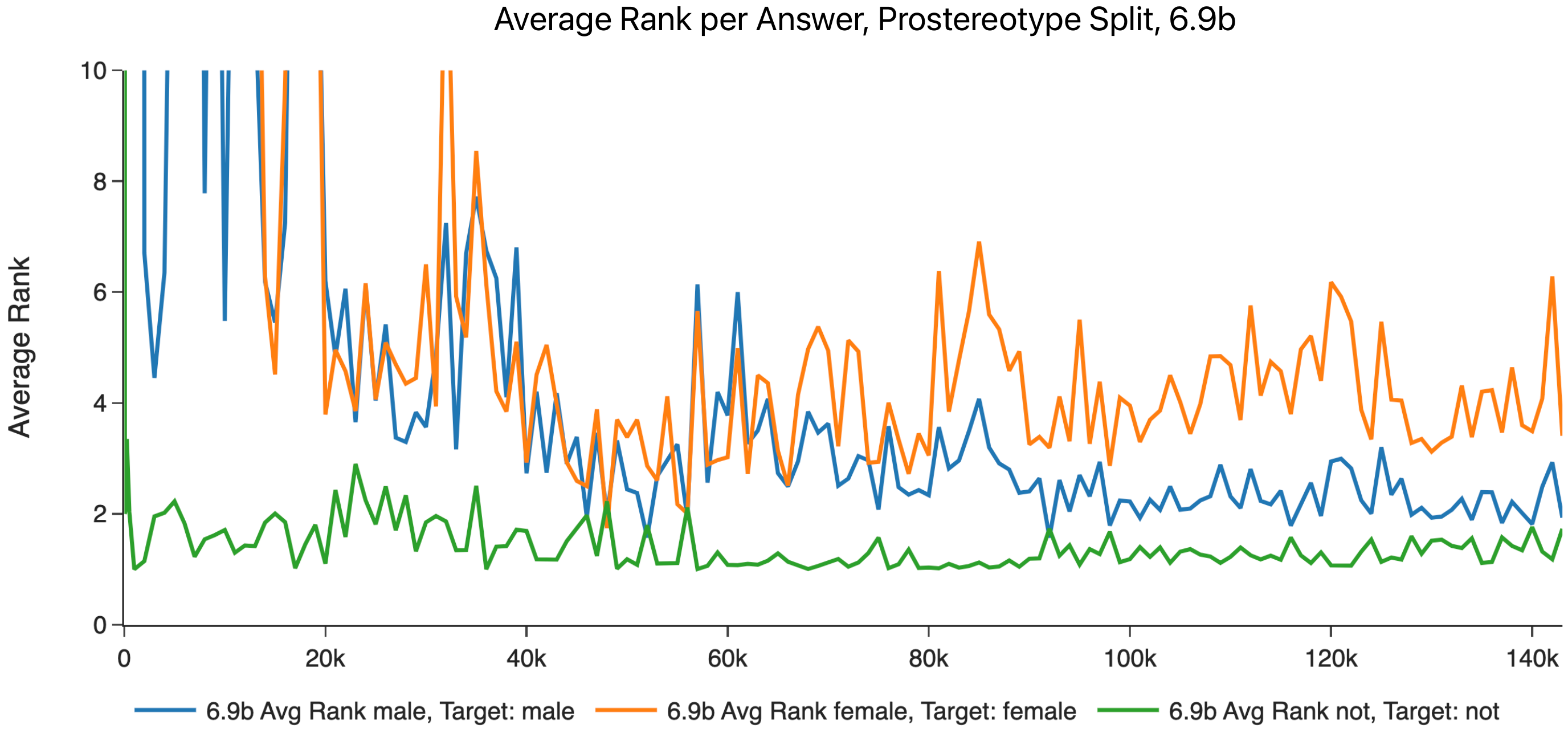}
    \caption{Average Rank per correct answer\\(``male'',``female'', ``not specified'') on Pythia-6.9b.}
    \label{fig:rank_results}
    \end{subfigure}%
  }\cr
  \noalign{\hfill}
  \hbox{%
    \begin{subfigure}{.5\textwidth}
    \centering
    \includegraphics[height=3.5cm,width=\textwidth]{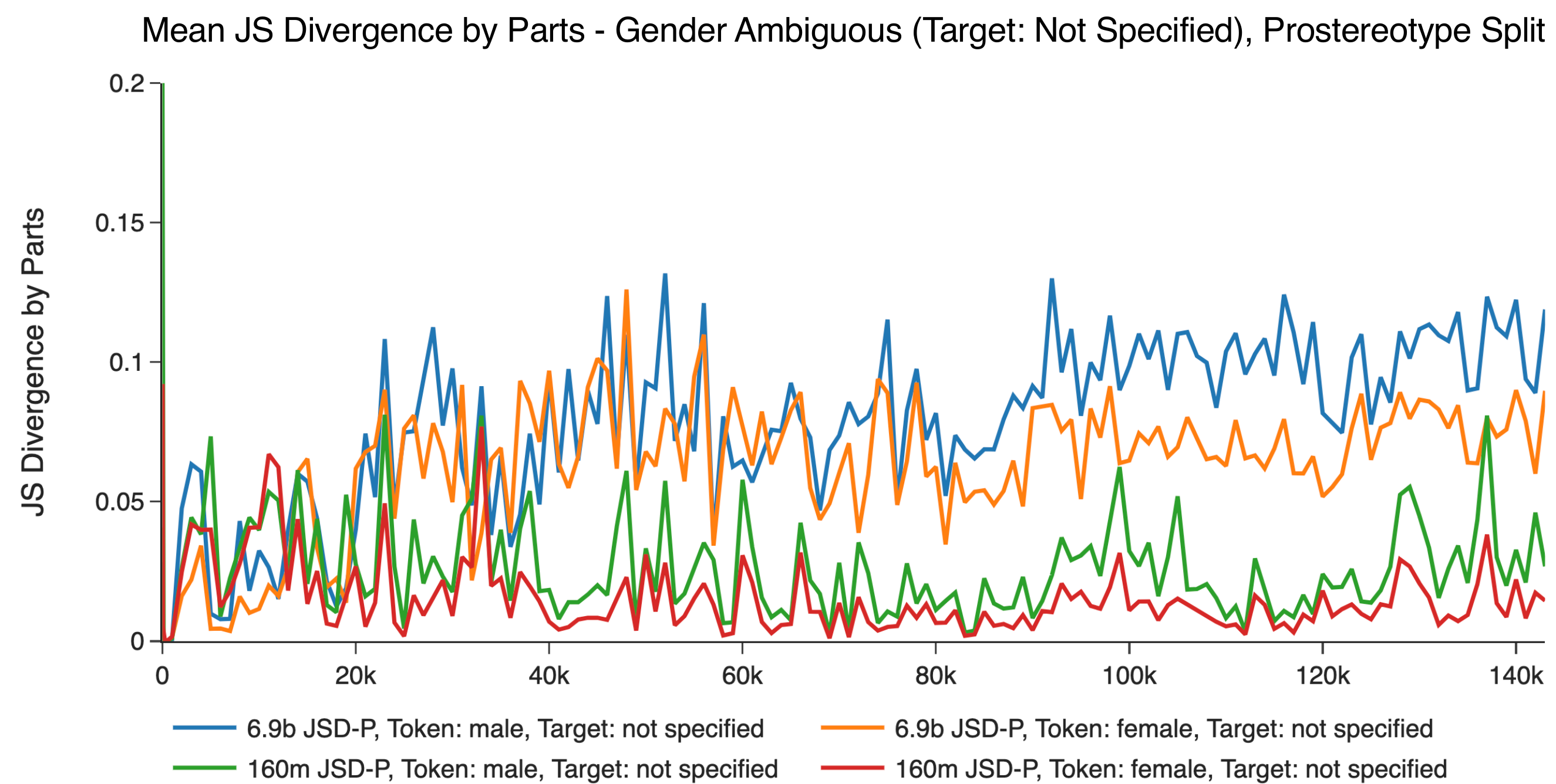}
    \caption{JSD-P on gendered options for prompts where ``not specified'' is correct, on Pythia 6.9b vs 160m.}
    \label{fig:kl_results}
    \end{subfigure}%
  }\cr
}

\caption{Prompting setup and select Pythia evaluation results. Further details in Figs.~\ref{fig:stopping_std},~\ref{fig:rank_std},~\ref{fig:not_std}.}
\vspace{-6mm}
\end{figure}

Each WinoBias sample has a stereotypically female occupation, a stereotypically male occupation, and a gendered pronoun referring to one of the subjects (Fig.~\ref{fig:prompt_example}); we use WinoBias Type 2 samples, where the pronoun unambiguously refers to one occupation. We generate two model prompts for each sample; one prompt queries the model on the gender of the occupation referred to by the pronoun, and the other queries the gender of the occupation not referenced by the pronoun (Fig.~\ref{fig:prompt_example}). Each prompt has three possible options (\textit{male}, \textit{female}, and \textit{not specified}), and only one answer.

We evaluate fairness dynamics during training with metrics that use next token probabilities: Average Rank (AR) for performance, and Jensen-Shannon Divergence by Parts (JSD-P) for fairness. AR computes the mean rank of the answer token's probability among the output probabilities for the entire vocabulary; lower rank indicates better performance. AR is more nuanced than accuracy, accounting for the magnitude of the error and not just its occurrence, enabling deeper insights into a model's poor performance (Fig.~\ref{fig:acc_ar}). For each answer option, JSD-P computes fairness as the divergence of the output token probability from the ideal one-hot categorical distribution; JS Divergence~\cite{61115} sums over all answer options whereas JSD-P is computed per answer option (App.~\ref{sec:kldc}). Smaller differences between each option's JSD-P is fairer and lower values are more confidently correct. JSD-P overcomes limitations in all-or-nothing-fairness metrics~\cite{biderman2023pythia, nadeem2020stereoset} by measuring both fairness and certainty to quantify bias (Fig.~\ref{fig:stereo_acc}). %This is crucial for text generation as sampling can amplify bias if probabilities are distributed like \{0.03, \textbf{0.95}, 0.02\}. 
This is crucial for text generation using sampling, since a distribution like \{0.03, \textbf{0.95}, 0.02\} would exhibit more bias than \{0.33, \textbf{0.34}, 0.32\}.

%AR contextualizes JSD-P with whether the model would generate the answer options, showing whether the bias between options measured by JSD-P is meaningful in practice.

% Our metrics are more nuanced than existing metrics, and we demonstrate that we are able to capture more detailed signals on performance, fairness, and uncertainty (see Appendix ~\ref{sec:metrics}).

% TODO add flexibility of how JSD-P can also be recombined into JS Divergence in Appendix, lower values of JSD, lower values are more confident / more correct dist of probability mass 

By tracking AR and JSD-P during training, in Figs.~\ref{fig:early_stopping},~\ref{fig:rank_results}, we establish that Pythia-6.9b can benefit from early stopping at $\approx80$k steps, trading a 1.7\% accuracy decrease on LAMBADA~\cite{paperno2016lambada} for a 92.5\% increase in fairness (mean JSD-P difference drops from 0.73 to 0.05). Further, in Fig.~\ref{fig:rank_results}, the AR for tokens ``male'' and ``female'' (when each is the correct answer) diverges after $\approx80$k steps; ``male'' AR improves, indicating a bias in performance towards ``male.'' %This indicates that as training progresses, a performance disparity builds between ``male'' and ``female,'' showing a bias in performance towards ``male.'' 
Using JSD-P, for Pythia-6.9b, we find larger probability mass on gendered answers, more so for ``male'' than ``female,'' for gender ambiguous prompts compared to Pythia-160m, showing that Pythia-6.9b is more likely to assume gender where it is unmentioned (Fig.~\ref{fig:kl_results}). Our results are significant under the Mann-Whitney U Test~\cite{mann1947test} with $p < 0.01$, rejecting the null hypothesis that the samples' underlying distributions are the same (Figs.~\ref{fig:mw_1},~\ref{fig:mw_2},~\ref{fig:mw_3}).

%Training progress reveals an increase in performance bias towards ``male'' increases, and an increase in gender assumption in Pythia-6.9b, unlike Pythia-160m (~\ref{fig:kl_results}). 

%In addition, we see that the rate of probability mass growth is higher for ``male'' than ``female''.
% ~\vspace{-5mm}
% ~\vspace*{-8mm}

\textbf{Summary}: We introduce AR and JSD-P to effectively characterize fairness dynamics during training, enabling the findings that: (1) common performance measures do not always reflect fairness during training, (2) early stopping can result in fairer models, (3) Pythia-6.9b is biased towards men, and (4) larger models can exhibit more bias in gender neutral contexts. Tracking fairness dynamics with our metrics can enable insights into bias development and 
opportunities for mitigation.

%to understand the advent of performance bias, and identify that larger models tend to make more biased assumptions when gender isn't specified. Using these metrics, 

% Taken together, we can understand the relationship between a model's performance and fairness, as well as identify growing disparities in performance across gender prediction and the tendency to assume gender where it is not specified.

% 1. Interrogating or critiquing the theoretical basis of existing evaluations
% 2. Novel methodologies for evaluating social impact across different AI modalities
%     1. Paradigm shift — look at during training too, not just fairness of end model 
% 3. Comparative analyses of existing evaluation frameworks and their effectiveness
%     * Mine and theirs

{
\small

\bibliography{bibliography}
\bibliographystyle{plain}

}

%%%%%%%%%%%%%%%%%%%%%%%%%%%%%%%%%%%%%%%%%%%%%%%%%%%%%%%%%%%%
\newpage
\appendix
\section{Limitations and Social Considerations}
Our evaluation is limited to the WinoBias dataset and the Pythia model family. WinoBias only examines bias across binary gender, which is a simplification of the contemporary understanding of gender~\cite{10.1145/3531146.3534627}. Further, WinoBias was constructed by referencing the US Bureau of Labor Statistics' data, meaning that the stereotypes evaluated are more reflective of the US and the Western world, and likely are not universal. In addition, the Pythia model family was trained for research purposes and not for use in production, so our results may not completely reflect how models trained for deployment behave. Lastly, when we suggest early stopping as a fairness intervention for Pythia-6.9b, we are only evaluating fairness on one axis (binary gender), so early stopping at the point identified may have some unintended consequences on other axes of bias. Further, early stopping simply works around bias, instead of truly mitigating it.

\section{JS Divergence by Parts}\label{sec:kldc}

JSD-P is similar to Jensen-Shannon Divergence (JS Divergence), but we do not sum across all individual divergence components. Instead, we examine each token's contribution to the overall divergence individually, in order to understand if certain answer options contribute to the overall divergence more than others. Therefore, JSD-P is more interpretable than JS Divergence.

We compute JSD-P individually across all potential answer tokens in $S$ (for our evaluation $S = \{male, female, not\}$) over a subset of prompts evaluated $W$ (in Fig.~\ref{fig:kl_results}, it is all prompts where the answer is ``not specified'').

Average JSD-P is computed using:
\begin{equation}
\text{$D(A(i)_{j}, B(i)_{j})$}_{i\in S, j\in W} = A(i)_{j} * log \Biggl( \frac{A(i)_{j}}{B(i)_{j}}\Biggr)
\end{equation}
\begin{equation}
\text{JSD-P}_{i\in S} = \frac{\sum_{j\in W} \frac{1}{2}\Biggl(D(P_{ideal}(i)_{j}, M(i)_{j}) + D(P(i)_{j},M(i)_{j})\Biggr)}{|W|} 
\label{eq:kldc}
\end{equation}
where:

\begin{eqexpl}[25mm]
\item{$P_{ideal}(x)_{y}$}$\begin{cases}
\begin{array}{ll}
0 & \text {if } x \textnormal{ is not correct answer}\\
1 & \text {if } x \textnormal{ is correct answer}\end{array}\end{cases}$ for token $x\in S$ and model prompt $y$
\smallskip
\item{$P(x)_{y}$}$softmax([\phi(male)_{y}, \phi(female)_{y}, \phi(not)_{y}])$ for token $x\in S$ and model prompt $y$, where $\phi(x)_{y}$ represents model output scores for prompt $y$
\smallskip
\item{$M(x)_{y}$} $\frac{1}{2}*(P(x)_y + P_{ideal}(x)_y)$ for token $x\in S$ and model prompt $y$
\end{eqexpl}

JSD-P measures the divergence between the model's output probabilities for each answer option and the correct answer (a one-hot categorical distribution). Differences in JSD-P between groups (like ``male'' or ``female'') indicate bias and unfair performance.

We utilize model outputs for ``not'' instead of combining the two token outputs for ``not specified,'' because in this context, for the Pythia models, ``specified'' follows ``not'' with high probability the majority of the time.

\section{Metrics Comparison}\label{sec:metrics}

\subsection{Average Rank vs Accuracy}

\begin{figure}[h]
    \centering
    \includegraphics[width=\linewidth]{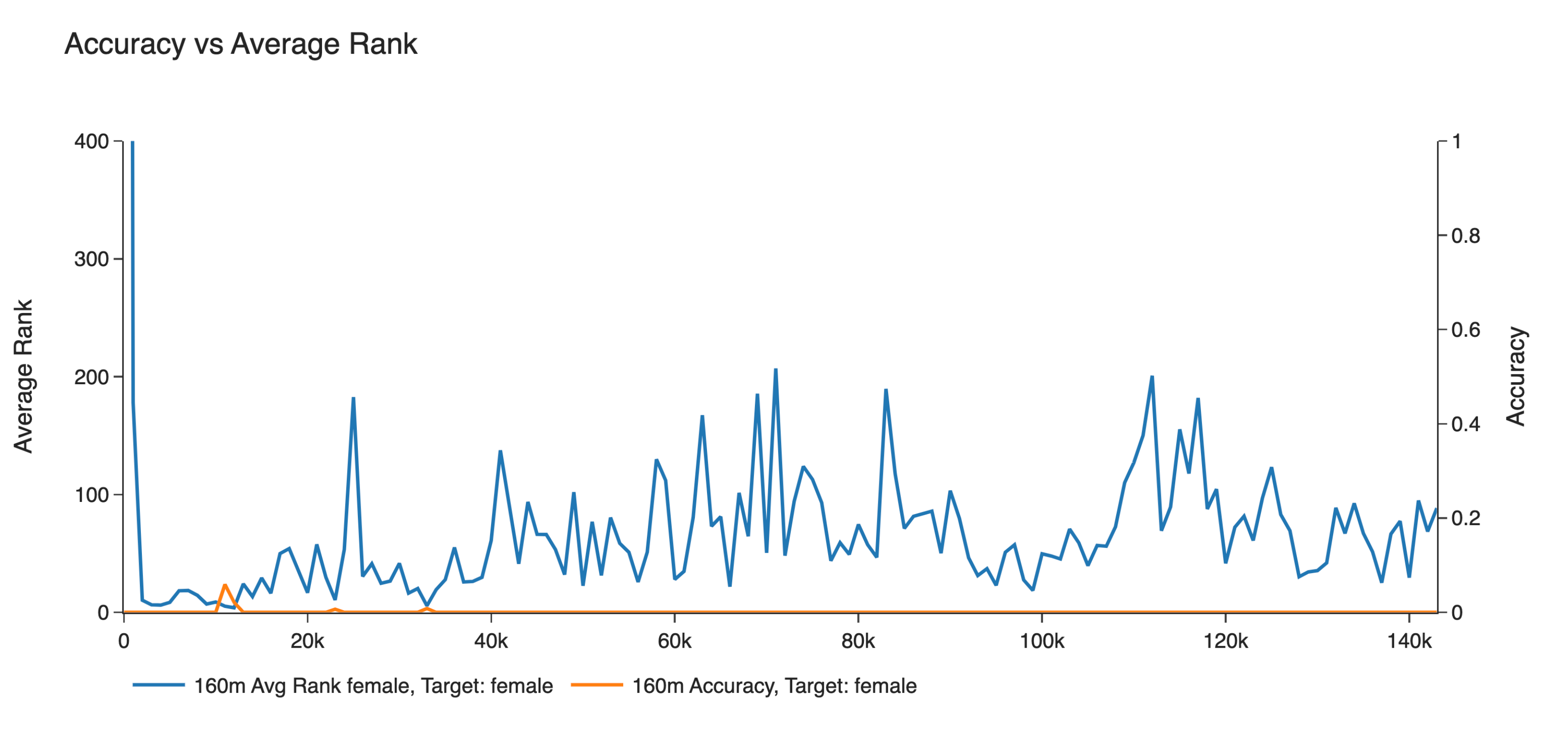}
    \caption{Average Rank for ``female'' captures more information than accuracy when the target answer is ``female'' for Pythia-160m. Accuracy remains close to 0 throughout training, while AR increases until $\approx$85k steps, declines between $\approx$85k and $\approx$100k steps, then increases again. }
    \label{fig:acc_ar}
\end{figure}

AR is particularly beneficial when examining poorly performing models. In Fig.~\ref{fig:acc_ar}, we can see that while accuracy is close to 0\% for the majority of training, AR increases until $\approx$85k steps, declines between $\approx$85k and $\approx$100k steps, then increases again. When selecting the most performant model checkpoint, AR indicates that we should select a model around step $\approx$100k, while accuracy cannot capture a difference in performance between any of these checkpoints. Further, since accuracy is a non-linear metric, when evaluated during training, it can lead to fallacies like observing the sudden emergence of good performance at a training step, when the emergence is simply due to the all-or-nothing nature of the metric \cite{schaeffer2023emergentabilitieslargelanguage}. The same issue does not hold for AR.

\subsection{JSD-P vs Stereotype Accuracy}

\begin{figure}[h]
    \centering
    \includegraphics[width=\linewidth]{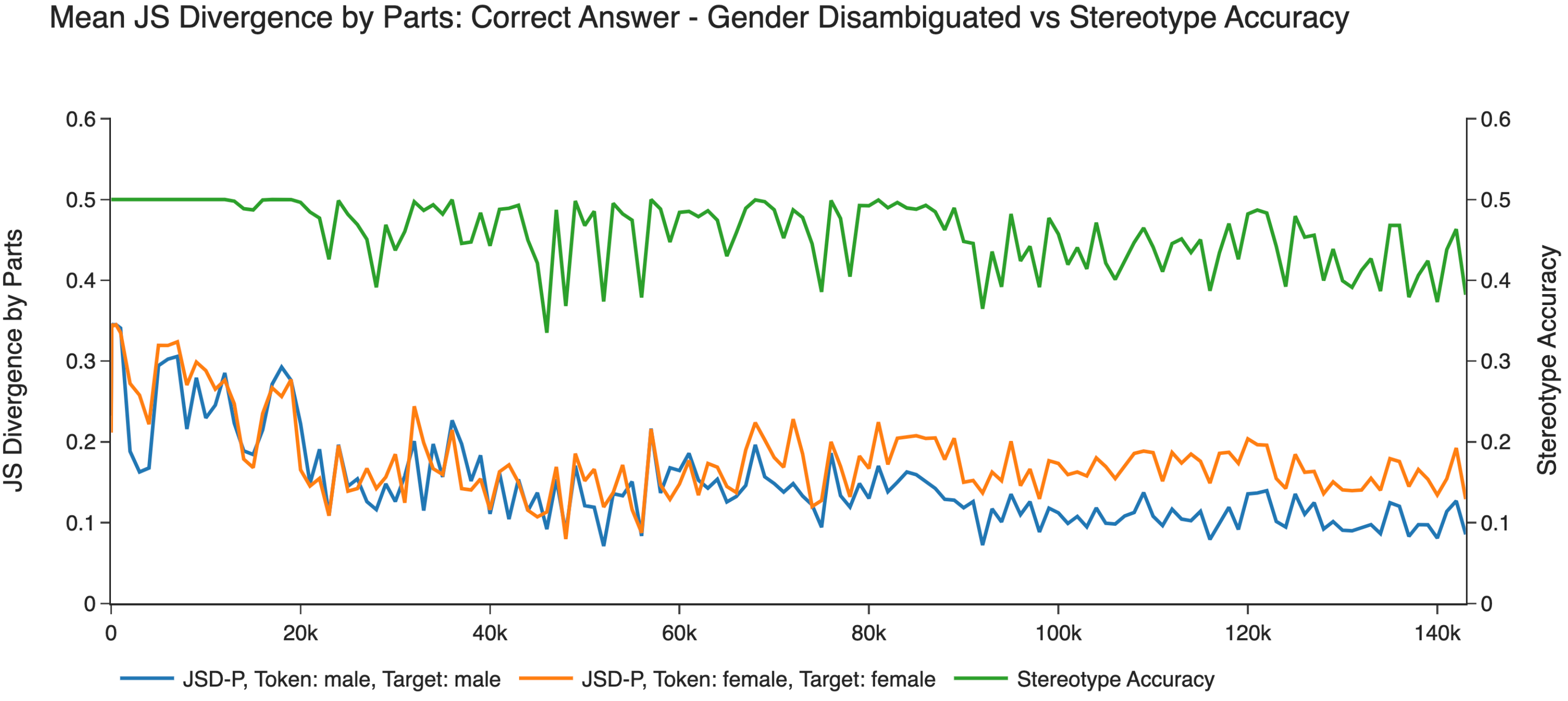}
    \caption{Stereotype Accuracy as defined by Biderman et al.~\cite{biderman2023pythia} vs JSD-P per correct answer (“male”, “female”). JSD-P per correct answer captures more information than Stereotype Accuracy. After $\approx$80k steps, there is a noticeable trend change in JSD-P per correct answer, while any change in Stereotype Accuracy is undetectable.}
    \label{fig:stereo_acc}
\end{figure}

We compare JSD-P with an all-or-nothing fairness metric called Stereotype Accuracy (SA) as defined in Biderman et al.~\cite{biderman2023pythia}. SA examines how accurately the model predicts stereotypical answers on the pro-stereotypical split of WinoBias. SA scores $1$ and $0$ are most biased, and $0.5$ is least biased (considered random). In Fig.~\ref{fig:stereo_acc}, SA slightly decreases throughout training, which indicates that the model's bias is slightly increasing during training. However, this lacks details found in Fig.~\ref{fig:early_stopping}; with JSD-P, we are able to understand that the model's confidence in its predictions increases over time, and more so when predicting ``male'' than ``female.''

% \begin{figure}
%     \centering
%     \includegraphics[width=\linewidth]{stereo_acc.pdf}
%     \caption{Stereotype Accuracy as defined by~\cite{biderman2023pythia}. Stereotype Accuracy slowly decreases as training progresses.}
%     \label{fig:stereo_acc}
% \end{figure}

\section{Standard Deviation and Statistical Significance}

\begin{figure}
\valign{#\cr
    \hbox{%
    \begin{subfigure}{\textwidth}
    \centering
    \includegraphics[width=\textwidth]{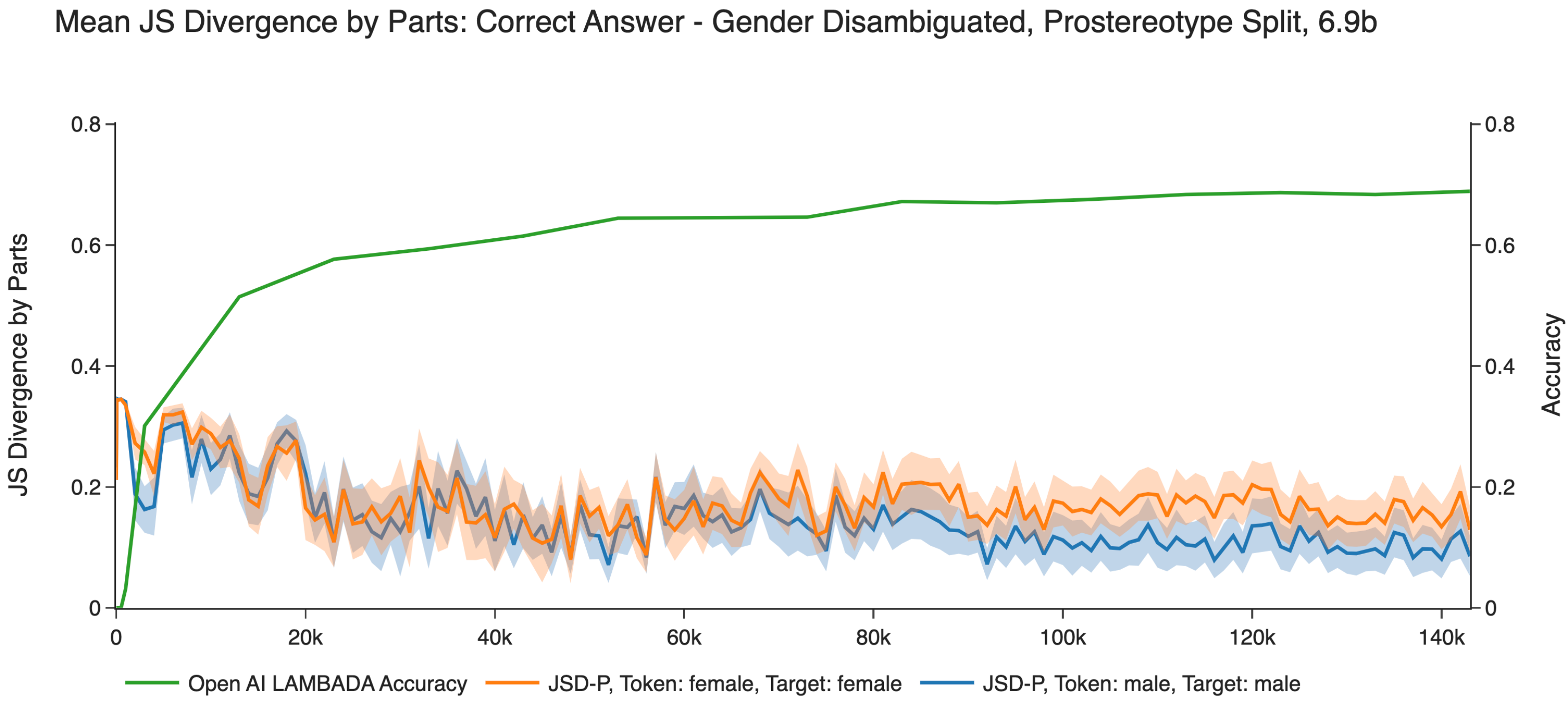}
    \caption{JSD-P per correct answer (``male'', ``female'') and performance on Open AI's LAMBADA benchmark on Pythia-6.9b, with standard deviations.}
    \label{fig:stopping_std}
    \end{subfigure}%
  }\cr
    \noalign{\hfill}
  \hbox{%
    \begin{subfigure}{\textwidth}
    \centering
    \includegraphics[width=\textwidth]{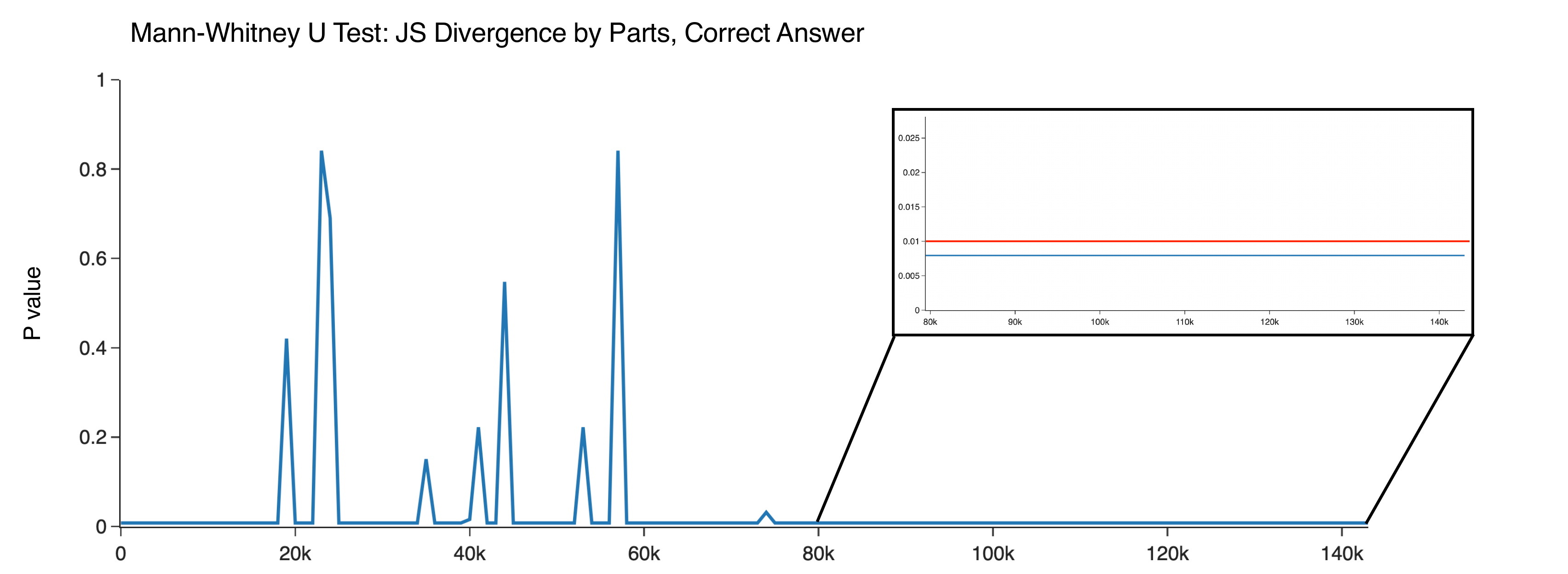}
    \caption{Mann-Whitney U Test illustrating that the JSD-P of ``male'' and ``female'' in Fig~\ref{fig:stopping_std} are significantly different $(p < 0.01)$ after $\approx$80k steps.}
    \label{fig:mw_1}
    \end{subfigure}%
  }\cr
  }
\caption{Detailed standard deviation and significance measures for Fig.~\ref{fig:early_stopping}.}

\end{figure}

\begin{figure}
\valign{#\cr
    \hbox{%
    \begin{subfigure}{\textwidth}
    \centering
    \includegraphics[width=\textwidth]{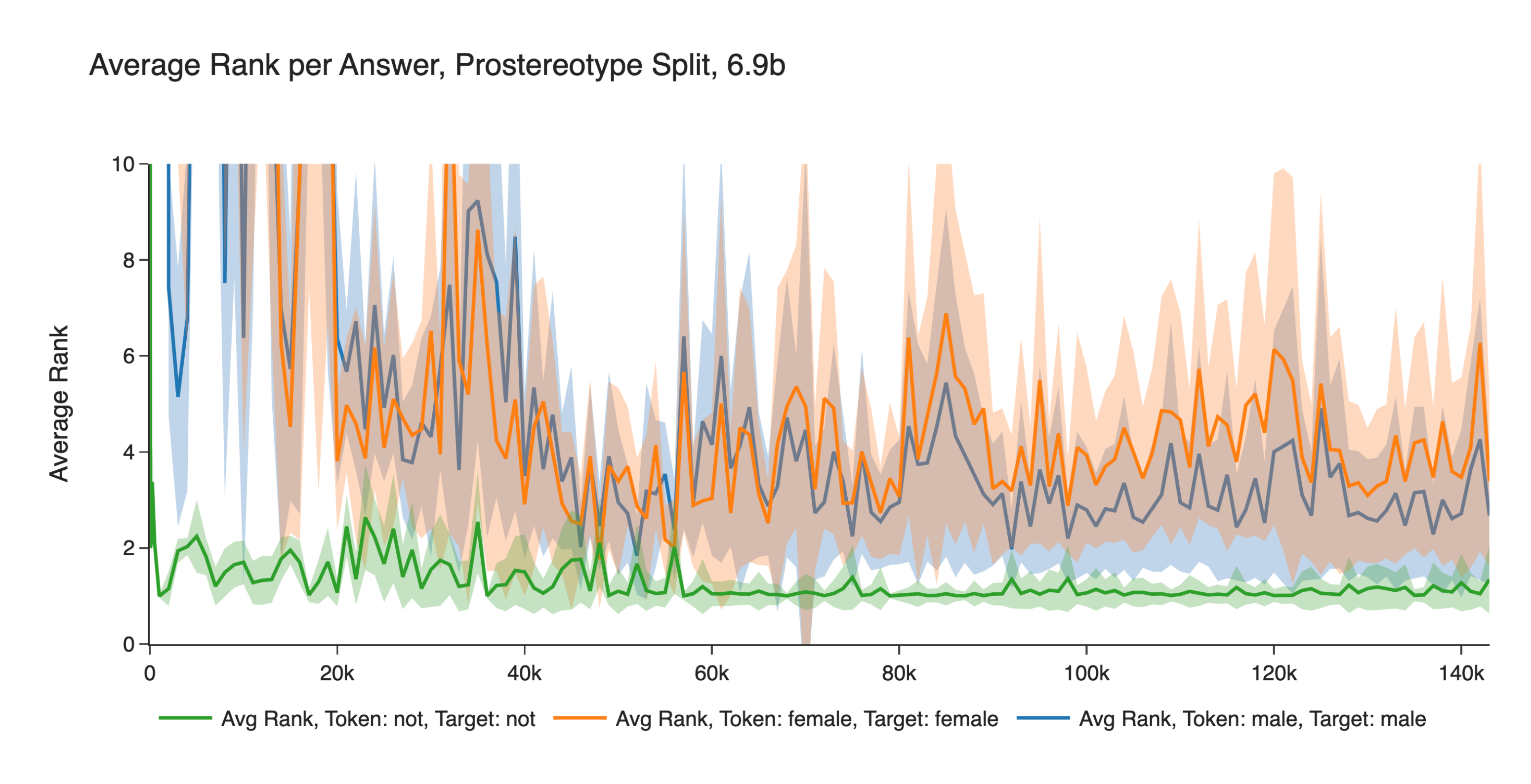}
    \caption{Average Rank per correct answer (``male'', ``female'', ``not specified'') on Pythia-6.9b, with standard deviations.}
    \label{fig:rank_std}
    \end{subfigure}%
  }\cr
    \noalign{\hfill}
  \hbox{%
    \begin{subfigure}{\textwidth}
    \centering
    \includegraphics[width=\textwidth]{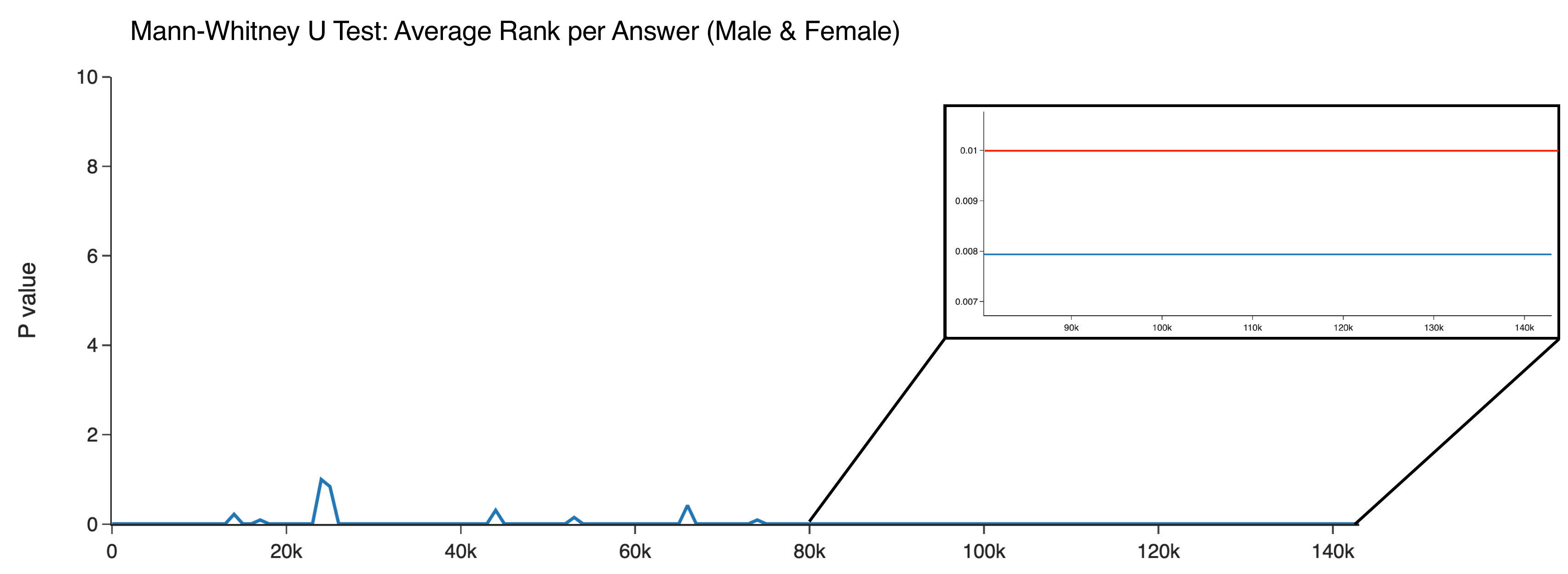}
    \caption{Mann-Whitney U Test illustrating that the AR of ``male'' and ``female'' in Fig~\ref{fig:rank_std} are significantly different $(p < 0.01)$ after $\approx$80k steps.}
    \label{fig:mw_2}
    \end{subfigure}%
  }\cr
  }
\caption{Detailed standard deviation and significance measures for Fig.~\ref{fig:rank_results}.}

\end{figure}

\begin{figure}
\valign{#\cr
    \hbox{%
    \begin{subfigure}{\textwidth}
    \centering
    \includegraphics[width=\textwidth]{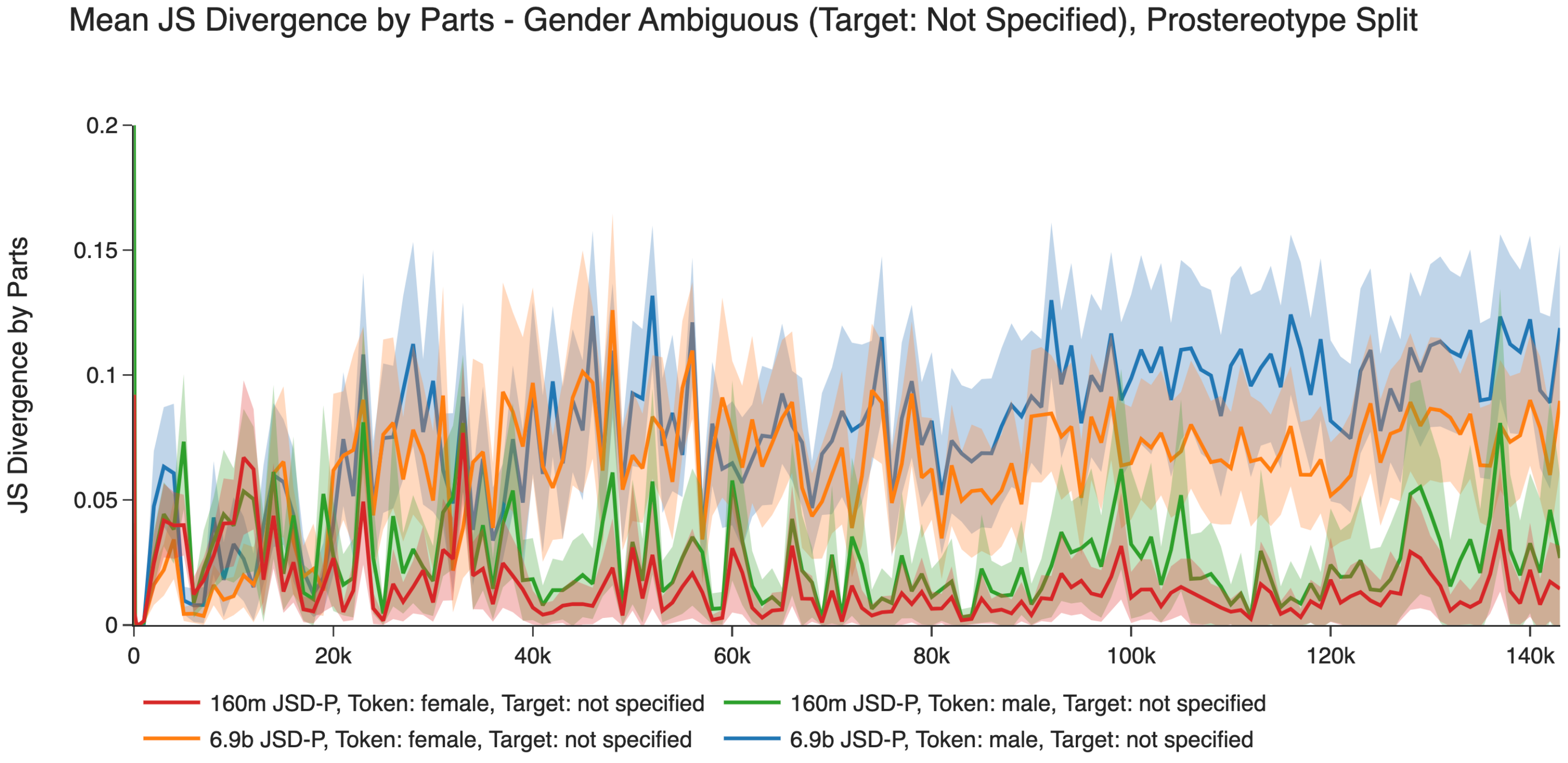}
    \caption{JSD-P on gendered options for prompts where ``not specified'' is correct on Pythia-6.9b vs 160m, with standard deviations.}
    \label{fig:not_std}
    \end{subfigure}%
  }\cr
    \noalign{\hfill}
  \hbox{%
    \begin{subfigure}{\textwidth}
    \centering
    \includegraphics[width=\textwidth]{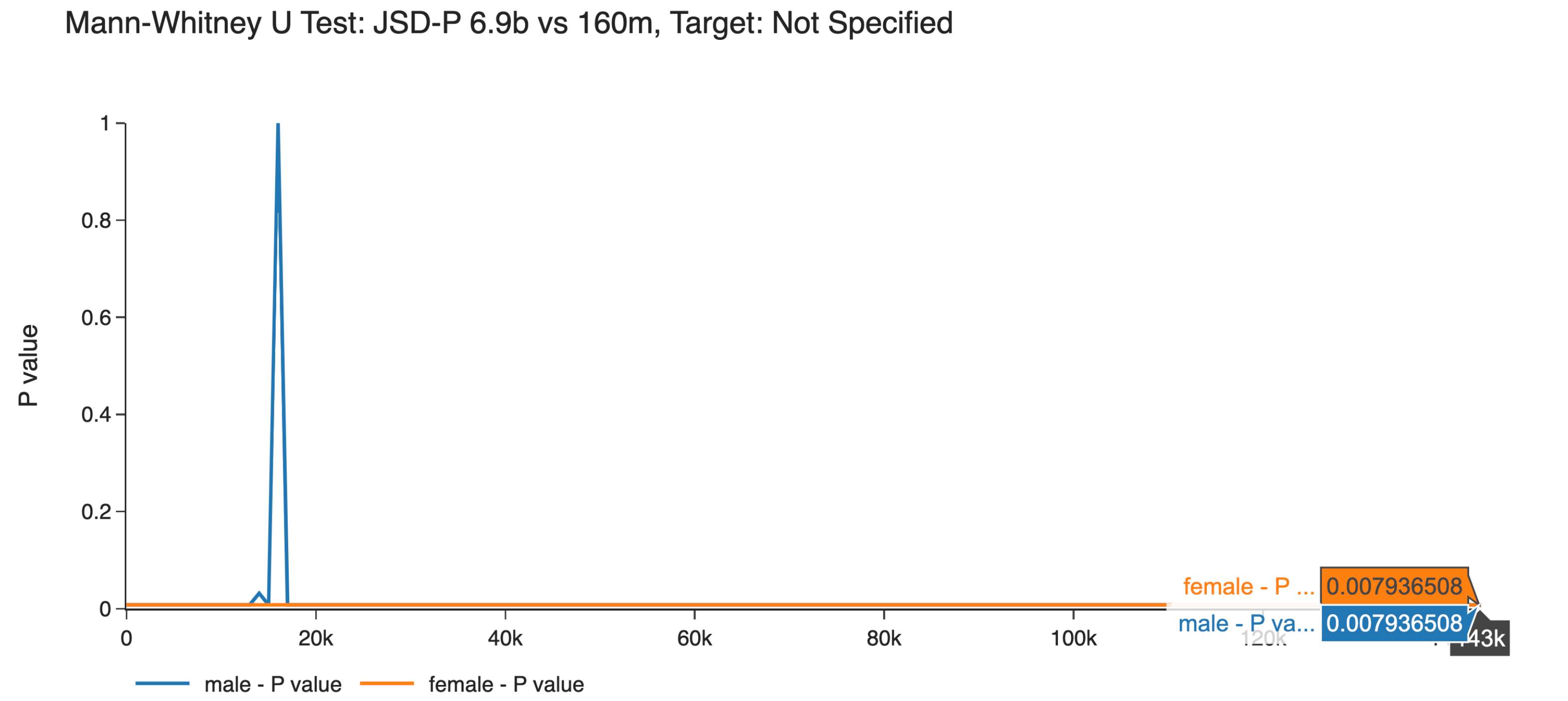}
    \caption{Mann-Whitney U Test illustrating that the JSD-P of ``male'' and ``female'' between Pythia-6.9b and 160m in Fig~\ref{fig:kl_results} are significantly different $(p < 0.01)$ throughout training.}
    \label{fig:mw_3}
    \end{subfigure}%
  }\cr
  }
\caption{Detailed standard deviation and significance measures for Fig.~\ref{fig:kl_results}.}

\end{figure}

Each experiment was repeated with 5 separate random seeds that determined the order of the options (``male,'' ``female,'' and ``not specified'') presented in each model prompt. For the figures presented in the main body, Figs.~\ref{fig:stopping_std},~\ref{fig:rank_std}, and~\ref{fig:not_std} calculate the standard deviation across these 5 runs, while Figs.~\ref{fig:mw_1},~\ref{fig:mw_2}, and~\ref{fig:mw_3} illustrate the significance of our results.

\end{document}